\documentclass[sn-mathphys-num]{sn-jnl}
\usepackage{graphicx}%
\usepackage{multirow}%
\usepackage{amsmath,amssymb,amsfonts}%
\usepackage{amsthm}%
\usepackage{mathrsfs}%
\usepackage[title]{appendix}%
\usepackage{xcolor}%
\usepackage{textcomp}%
\usepackage{manyfoot}%
\usepackage{booktabs}%
\usepackage{algorithm}%
\usepackage{algorithmicx}%
\usepackage{algpseudocode}%
\usepackage{listings,float}%
\usepackage{xcolor}
\newcommand{\revise}[1]{\textcolor{black}{#1}}
\usepackage{listings}
\usepackage{color}

\definecolor{lightgray}{rgb}{0.95,0.95,0.95}

\lstdefinestyle{promptstyle}{
    backgroundcolor=\color{lightgray!50},
    basicstyle=\ttfamily\small\color{black},
    breaklines=true,
    breakindent=0pt,
    frame=single,
    framesep=5pt,
    numbers=none,
    prebreak={},
    postbreak={},
    xleftmargin=5pt,
    xrightmargin=5pt,
    belowskip=10pt,
    aboveskip=10pt,
    columns=fullflexible,
    keepspaces=true
}
\begin{document}

\title[Article Title]{A Hybrid Framework with Large Language Models for Rare Disease Phenotyping}


\author*[1,6]{\fnm{Jinge} \sur{Wu}}\email{jinge.wu.20@ucl.ac.uk}

\author[2]{\fnm{Hang} \sur{Dong}}\email{H.Dong2@exeter.ac.uk}

\author[3]{\fnm{Zexi} \sur{Li}}\email{zexi.li@stcatz.ox.ac.uk}

\author[4]{\fnm{Haowei} \sur{Wang}}\email{rmhahw2@ucl.ac.uk}

\author[5]{\fnm{Runci} \sur{Li}}\email{runci.li.20@ucl.ac.uk}

\author*[6]{\fnm{Arijit} \sur{Patra}}\email{arijit.patra@ucb.com}

\author[6]{\fnm{Chengliang} \sur{Dai}}\email{chengliang.dai@ucb.com}

\author[6]{\fnm{Waqar} \sur{Ali}}\email{waqar.ali@ucb.com}

\author[6]{\fnm{Phil} \sur{Scordis}}\email{phil.scordis@ucb.com}

\author*[1,7]{\fnm{Honghan} \sur{Wu}}\email{honghan.wu@ucl.ac.uk}
\affil[1]{\orgdiv{Institute of Health Informatics}, \orgname{University College London}, \orgaddress{\city{London}, \country{United Kingdom}}}

\affil[2]{\orgdiv{Department of Computer Science}, \orgname{University of Exeter}, \orgaddress{\city{Exeter}, \country{United Kingdom}}}

\affil[3]{\orgdiv{The Nuffield Department of Surgical Sciences}, \orgname{University of Oxford}, \orgaddress{\city{Oxford}, \country{United Kingdom}}}

\affil[4]{\orgdiv{Division of Medicine}, \orgname{University College London}, \orgaddress{\city{London}, \country{United Kingdom}}}

\affil[5]{\orgdiv{EGA- Institute for Women‘s Health}, \orgname{University College London}, \orgaddress{\city{London}, \country{United Kingdom}}}

\affil[6]{\orgname{UCB Pharma UK}, \orgaddress{\city{Slough}, \country{United Kingdom}}}

\affil[7]{\orgdiv{School of Health and Wellbeing}, \orgname{University of Glasgow}, \orgaddress{\city{Glasgow}, \country{United Kingdom}}}

\abstract{\textbf{Purpose:} Rare diseases pose significant challenges in diagnosis and treatment due to their low prevalence and heterogeneous clinical presentations. Unstructured clinical notes contain valuable information for identifying rare diseases, but manual curation is time-consuming and prone to subjectivity. This study aims to develop a hybrid approach combining dictionary-based natural language processing (NLP) tools with large language models (LLMs) to improve rare disease identification from unstructured clinical reports.\\
\textbf{Methods:} We propose a novel hybrid framework that integrates the Orphanet Rare Disease Ontology (ORDO) and the Unified Medical Language System (UMLS) to create a comprehensive rare disease vocabulary. SemEHR, a dictionary-based NLP tool, is employed to extract rare disease mentions from clinical notes. To refine the results and improve accuracy, we leverage various LLMs, including LLaMA3, Phi3-mini, and domain-specific models like OpenBioLLM and BioMistral. Different prompting strategies, such as zero-shot, few-shot, and knowledge-augmented generation, are explored to optimize the LLMs' performance.\\
\textbf{Results:} The proposed hybrid approach demonstrates superior performance compared to traditional NLP systems and standalone LLMs. LLaMA3 and Phi3-mini achieve the highest F1 scores in rare disease identification. Few-shot prompting with 1-3 examples yields the best results, while knowledge-augmented generation shows limited improvement. Notably, the approach uncovers a significant number of potential rare disease cases not documented in structured diagnostic records, highlighting its ability to identify previously unrecognized patients.\\
\textbf{Conclusion:} The hybrid approach combining dictionary-based NLP tools with LLMs shows great promise for improving rare disease identification from unstructured clinical reports. By leveraging the strengths of both techniques, the method demonstrates superior performance and the potential to uncover hidden rare disease cases. Further research is needed to address limitations related to ontology mapping and overlapping case identification, and to integrate the approach into clinical practice for early diagnosis and improved patient outcomes.}




\maketitle

\section{Introduction}\label{sec1}

Rare diseases, defined as those affecting less than 200,000 individuals in the United States and fewer than 1 in 2,000 people in Europe, pose significant challenges to patients, healthcare systems, and society at large \cite{groft2021progress}. These conditions, often chronic and life-threatening, collectively impact millions of people worldwide, with an estimated 300 million individuals living with a rare disease globally. The low prevalence and diverse clinical manifestations of rare diseases create substantial hurdles in diagnosis, treatment, and research efforts \cite{schieppati2008rare}. 

The journey to diagnosis for rare disease patients is often long and arduous, with many experiencing a \revise{``}diagnostic odyssey" that can last years or even decades \cite{bauskis2022diagnostic}. This delay in diagnosis has profound consequences for patients, including inappropriate treatments, unnecessary medical procedures, and missed opportunities for early intervention \cite{hampson2022measuring}. The emotional and psychological toll on patients and their families is immense, as they struggle with uncertainty, isolation, and the challenges of navigating complex healthcare systems ill-equipped to address their unique needs \cite{stoller2018challenge,thygesen2023nationwide}.

Underdiagnosis of rare diseases not only impacts individual patients but also places a significant burden on healthcare systems \cite{zhang2023diagnosing}. Misdiagnoses and delayed diagnoses lead to increased healthcare utilization, with patients often consulting multiple specialists and undergoing numerous tests before receiving an accurate diagnosis. This inefficiency strains healthcare resources and contributes to escalating costs. Moreover, the absence of timely and accurate diagnoses hinders the development of targeted therapies and limits patients' access to appropriate care and support services\cite{bauskis2022diagnostic,griggs2009clinical}.

The challenges posed by rare diseases extend beyond clinical care to research and drug development. The scarcity of patients with a positive diagnosis for a given rare disease complicates the conduct of clinical trials and the collection of robust epidemiological data \cite{zhang2023diagnosing}. This, in turn, hampers efforts to understand disease mechanisms, identify potential therapeutic targets, and develop effective treatments. The underdiagnosis of rare diseases further exacerbates this issue by limiting the pool of identified patients who could participate in research studies or clinical trials \cite{bauskis2022diagnostic}.

In light of these challenges, there is an urgent need for innovative approaches to improve rare disease identification and diagnosis \cite{arbabi2019identifying,cook2019guide,beaulieu2023predicting}. Unstructured clinical text data, such as clinical notes, offers a rich source of information for rare disease identification, but manual curation is laborious and subjective. Automated natural language processing (NLP) tools that can effectively extract symptoms or diagnoses from unstructured clinical text data play a crucial role in improving rare disease patient diagnosis, treatment, and research. These tools have the potential to revolutionize the field by enabling large-scale rare disease identification, facilitating better medical outcomes for these vulnerable patients in the healthcare system.

Traditional approaches to disease identification often rely on manual curation by domain experts, which is time-consuming, labor-intensive, and subject to human biases. To address these challenges, there has been a growing interest in developing computational methods for automated identification. Dictionary-based NLP tools have been widely used to extract structured information from unstructured clinical narratives, such as electronic health records (EHRs) and scientific literature \cite{wu2018semehr,dong2023ontology}. These tools leverage pre-defined rules and dictionaries to identify and normalize phenotypic concepts. However, dictionary-based systems often struggle with the complexity and variability of natural language, leading to suboptimal performance in capturing the nuances of rare disease phenotypes.

Recent advancements in large language models (LLMs), such as the GPT \cite{achiam2023gpt} and LLaMA \cite{touvron2023llama} series, have revolutionized the field of NLP. These models, pretrained on vast amounts of text data, have demonstrated remarkable capabilities in understanding and generating human-like language. By leveraging the knowledge embedded in LLMs, researchers can potentially enhance the performance of dictionary-based NLP tools for rare disease identification. However, LLMs are susceptible to biases present in their training data and can struggle with factual accuracy, particularly in specialized domains like medicine. Furthermore, their \revise{``}black box" nature makes it difficult to understand their reasoning and identify potential errors \cite{wang2023survey}.

To address these limitations, we propose a novel hybrid approach that combines the strengths of ontology and dictionary-based NLP tools with the capabilities of LLMs. This approach leverages the interpretability and control of dictionary-based systems to guide the LLM's analysis, potentially improving its accuracy and focus on identifying rare diseases within unstructured clinical data. Our main contributions can be concluded as follows:
\begin{itemize}
    \item We propose a novel hybrid framework that integrates ontology and dictionary-based NLP tools with fine-tuned LLMs to enhance the accuracy of rare disease identification from clinical notes. This approach leverages the strengths of both techniques: the high recall of dictionary-based systems guided by a comprehensive vocabulary derived from ORDO/UMLS and the contextual understanding of LLMs.
    
    \item To optimize contextual reasoning within the hybrid framework, we conduct extensive experiments with diverse LLMs, exploring various prompt methods (zero-shot, few-shot, knowledge-augmented generation) and context lengths. These experiments provide valuable insights into the impact of these factors on rare disease identification accuracy.
    
    \item We further apply our methods to the large scale real-world patient notes. Our analysis reveals a substantial number of potential rare disease cases that are not currently documented in structured diagnostic records. This finding highlights the immense potential of our method for uncovering hidden rare disease cases, facilitating early diagnosis, and ultimately improving patient outcomes and treatment development.
\end{itemize}


\section{Related Work}

\subsection{Rare Disease Identification as Text Phenotyping with Ontologies}

Cohort identification, the process of identifying cases of disease from clinical records \cite{ford2016extracting}, is a crucial task in healthcare research and clinical practice. This task is typically accomplished through the use of clinical codes, such as the International Classification of Diseases (ICD), or by analyzing unstructured data, such as clinical notes. When free text clinical notes are used as the primary source for cohort identification, the task is referred to as \textit{text phenotyping}. Text phenotyping involves extracting relevant information about patient phenotypes, including symptoms, signs, and diagnoses, from the unstructured text data.

Ontologies play a pivotal role in rare disease identification, as they provide structured and standardized information on rare diseases and their associated phenotypes. The key ontologies used in this domain include the Human Phenotype Ontology (HPO) \cite{robinson2008human}, Orphanet \cite{weinreich2008orphanet}, and the Online Mendelian Inheritance in Man (OMIM) \cite{hamosh2000online}. HPO is a standardized vocabulary of phenotypic abnormalities encountered in human disease, enabling the consistent description of phenotypic information across different databases and applications. Orphanet is a comprehensive resource for information on rare diseases and orphan drugs, providing a nomenclature and classification of rare diseases. OMIM is a compendium of human genes and genetic disorders, offering detailed information on the molecular basis of inherited diseases.
These specialized ontologies are often linked to more general ontologies, such as the Unified Medical Language System (UMLS) \cite{bodenreider2004unified} and the International Classification of Diseases, 10th Revision (ICD-10) \cite{world2004international}. UMLS is a comprehensive collection of biomedical vocabularies and standards, facilitating the integration of information from various sources. ICD-10 is a widely used system for classifying diseases, injuries, and health conditions, providing a standardized coding scheme for clinical and research purposes.

The integration of these ontologies allows for a more comprehensive and accurate representation of rare diseases and their associated phenotypes. By leveraging the structured information provided by these ontologies, researchers and clinicians can improve the efficiency and accuracy of rare disease identification from clinical texts.

\subsection{Natural Language Processing for Text Phenotyping}

Dictionary-based NLP tools, such as cTAKES \cite{savova2010mayo}, SemEHR \cite{wu2018semehr}, and MedCAT \cite{kraljevic2019medcat}, have been widely adopted to extract information from clinical narratives, including electronic health records (EHRs) and scientific literature. These tools utilize predefined rules and dictionaries to identify and normalize phenotypic concepts, transforming raw textual data into a structured format for further analysis. By leveraging extensive vocabularies and ontologies, such as UMLS and HPO, these tools can recognize a wide range of medical terms, abbreviations, and synonyms, facilitating accurate concept identification within the text.

However, dictionary-based NLP tools often struggle with the variability and ambiguity inherent in natural language, as the semantic meaning of phrases and sentences can be highly context-dependent. For instance, the phrase \revise{``}cold" can refer to a viral infection, a sensation of low temperature, or a personality trait, depending on the context. Dictionary-based tools, which primarily rely on predefined rules and dictionaries, may lack the ability to fully capture and disambiguate such semantic nuances.

Moreover, the performance of dictionary-based NLP tools can be limited by the comprehensiveness and quality of the underlying dictionaries and ontologies. While resources like UMLS and HPO provide extensive coverage of medical concepts, they may not always include the most up-to-date terminology or capture the full spectrum of rare disease phenotypes, leading to the omission of important information or misclassification of concepts, particularly in the context of rare diseases where the associated phenotypes may be atypical or poorly characterized.

Another challenge faced by dictionary-based NLP tools is the handling of negation and uncertainty, as clinical narratives often contain negated or hypothetical statements. These expressions can significantly alter the meaning of the associated concepts and require careful consideration during information extraction. Dictionary-based tools may struggle to accurately identify and interpret such negation and uncertainty, potentially leading to errors in the extracted information.

To address these limitations, researchers have explored various approaches to enhance the performance of dictionary-based NLP tools, including the incorporation of machine learning techniques. Machine learning and deep learning models have emerged as powerful tools for automating phenotyping from EHRs and other clinical data sources, with the ability to learn complex patterns and relationships from large volumes of data, enabling them to identify and extract relevant phenotypic information with minimal human intervention.

Traditional machine learning algorithms, such as logistic regression, support vector machines (SVMs), and random forests, have been successfully applied to phenotyping tasks \cite{kirby2016phekb, yu2018enabling}. These models can learn from labeled training data to classify patients based on the presence or absence of specific phenotypes, and have shown promising results in predicting the risk of developing certain conditions based on a combination of clinical and demographic features.

In recent years, deep learning models, particularly convolutional neural networks (CNNs) and recurrent neural networks (RNNs), have gained significant attention for their ability to learn hierarchical representations from complex and unstructured data \cite{gehrmann2018comparing}. CNNs are well-suited for processing grid-like data, such as medical images or two-dimensional representations of clinical notes, while RNNs are designed to handle sequential data, such as time series or natural language. These models have been applied to various phenotyping tasks, such as detecting abnormalities in medical images or identifying specific clinical events from EHR data, achieving high accuracy and outperforming traditional machine learning approaches \cite{tomašev2019clinically}.

More recently, pre-trained language models, such as BERT (Bidirectional Encoder Representations from Transformers), have shown remarkable performance in various natural language processing tasks, including phenotyping from unstructured clinical notes \cite{li2020behrt}. BERT-based models can capture the contextual information and long-range dependencies in the text, allowing for a more accurate understanding of the semantic meaning and relationships between clinical concepts, and have achieved state-of-the-art results in tasks such as named entity recognition, relation extraction, and document classification. \revise{BERT-based models can be effectively adapted by adding an extra layer fined-tuned for Named Entity Recognition using token-level labels, an example is BioBERT-NER \cite{alonso2021named} fine-tuned on top of BioBERT \cite{lee2020biobert}. A recent example of fine-tuning BERT for rare disease phenotyping is PhenoBERT \cite{feng2022phenobert}, matching text fragments to concepts in the HPO ontology, and a hierarchy-aware CNN-based model was used for narrowing the candidates from the large number of concepts in the ontology.}

However, despite the progress made by machine learning and deep learning models in phenotyping, their performance on rare disease phenotyping remains limited by several factors, particularly the scarcity of labeled training data for rare diseases. Due to the low prevalence of rare diseases, obtaining a large enough dataset with reliable labels can be extremely difficult and costly, hindering the ability of models to generalize and capture the complex phenotypic expressions of rare diseases.

Recently, Large language models (LLMs), such as the GPT \cite{achiam2023gpt} and LLaMA \cite{touvron2023llama} series, have revolutionized the field of natural language processing with their impressive performance across a wide range of tasks, including named entity recognition, relation extraction, and question answering \cite{lee2020biobert, rasmy2021med}. These models, pre-trained on vast amounts of textual data, have the ability to capture complex linguistic patterns and generate human-like responses, making them a promising tool for various applications in the biomedical domain.

\revise{LLMs like GPT or LLAMA can be prompted for named entity recognition and concept normalization, and this usually works after fine-tuning or instruction-tuning. The study \cite{shyr2024identifying} shows that ChatGPT-3.5 model cannot match the results of BERT-based approach for rare disease phenotype extraction. The work \cite{yang2024enhancing} explored the fine-tuning of GPT and LLAMA models for phenotype concept recognition from HPO ontology, using data annotations created with an NER tool and supported by manual experts; results showing their comparable performances to BERT-based fine-tuning. The study \cite{wang2024fine} uses concept names, identifiers, and synonyms in HPO ontology as data to fine-tune LLAMA 2 and greatly enhance the performance for concept normalisation, compared to ChatGPT-3.5. Both works provide useful models (PhenoGPT \cite{yang2024enhancing} and PhenoHPO \cite{wang2024fine}) for rare disease concept normalization to HPO ontology and they only use LLMs alone in the inference. We will compare both studies with our dictionary and LLM-based hybrid approach for rare disease identification.}

\revise{Two other related works use LLMs for textual data with rare diseases \cite{thompson2023large,oniani2023large}.} The work \cite{thompson2023large} applied LLMs to a specific rare disease \cite{thompson2023large}, and introduced a zero-shot LLM-based method enriched by retrieval-augmented generation and MapReduce for identifying pulmonary hypertension (PH) from clinical notes. By combining the strengths of LLMs with retrieval-based methods and distributed computing techniques like MapReduce, they were able to accurately identify PH cases without the need for labeled training data. \revise{The work} \cite{oniani2023large} investigated prompting strategies for text identification using LLMs. By carefully crafting prompts that target specific rare diseases or phenotypic characteristics, the authors were able to elicit accurate and informative responses from the LLMs. However, their work focused on the four most frequent rare diseases in the MIMIC-IV dataset, rather than the full set of thousands of rare diseases defined in an ontology, highlighting the need for further investigation into the scalability and generalizability of these approaches.

Our work aims to combine the dictionary-based method and LLM for a comprehensive rare disease phenotyping approach. By leveraging the strengths of both methods, we seek to address the limitations of each individual approach and develop a more accurate and robust system for identifying rare diseases from clinical narratives. This hybrid approach has the potential to improve the efficiency and effectiveness of rare disease phenotyping, ultimately leading to better patient care and research outcomes.

\section{Methods}

\begin{figure*}[h]
    \includegraphics[width=\columnwidth]{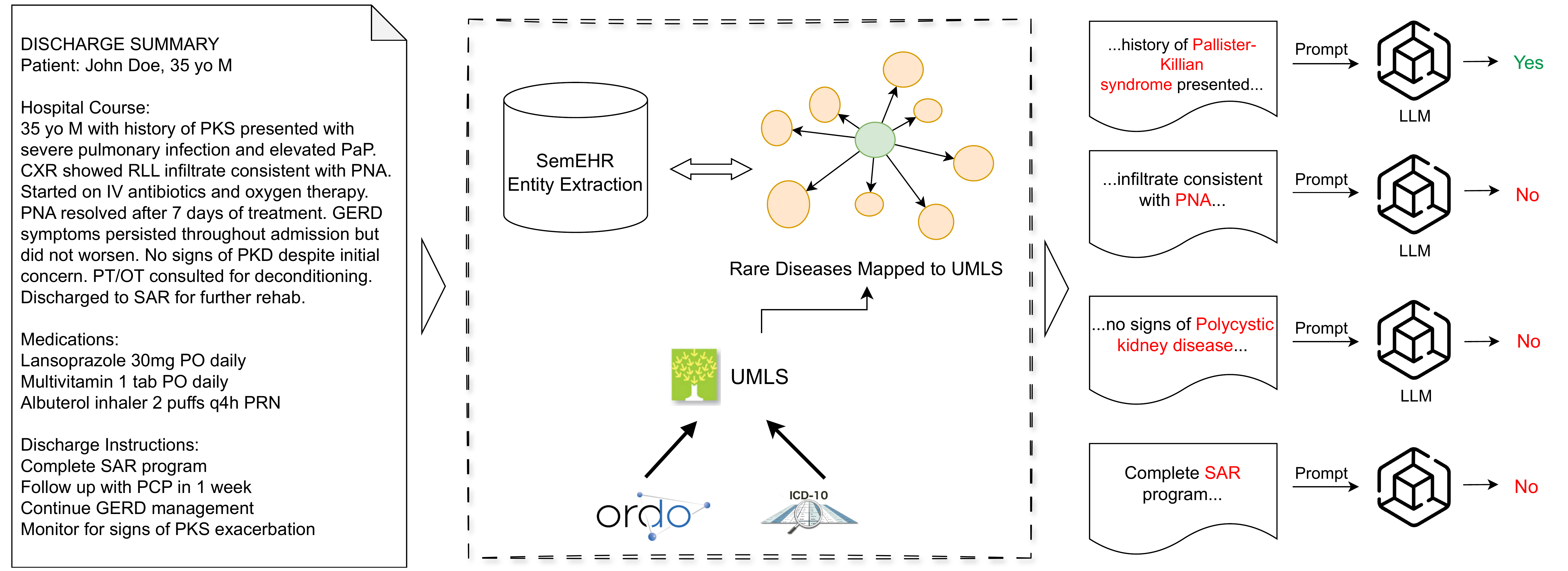}
  \caption{Overview of our work. In the given example report, there are some abbreviations that are mistakenly extracted as rare diseases (eg. PNA: Pneumonia, SAR: Subacute rehabilitation), and also some negated mentioned are extracted (eg. no signs of xxx). We propose leveraging LLMs for enhanced contextual filtering, enabling more precise determinations of relevance and validity within the extracted information.}
  \label{fig:overview}
\end{figure*}

Figure \ref{fig:overview} summarizes our overall work.

\subsection{Problem Statement}

Given a set of unstructured clinical notes $\mathcal{D} = {d_1, d_2, \ldots, d_N}$, and a comprehensive rare disease ontology $\mathcal{O}$ that defines a set of rare diseases $\mathcal{R} = {r_1, r_2, \ldots, r_K}$ along with their associated phenotypic characteristics $\mathcal{P} = {p_1, p_2, \ldots, p_L}$, the objective is to develop a hybrid approach that combines dictionary-based methods and LLMs to accurately identify and extract rare disease mentions and their corresponding phenotypic information from the clinical notes.
Let $f: \mathcal{D} \rightarrow \mathcal{R} \times \mathcal{P}$ be a function that maps each clinical note $d_i$ to a set of rare disease phenotypes ${r_j}$ or ${p_k}$, where $r_j \in \mathcal{R}$ and $p_k \in \mathcal{P}$. The goal is to optimize the function $f$ by leveraging the strengths of both dictionary-based methods and LLMs.

\subsection{Rare Disease Terminology from Ontology}
\label{section:RD_onto}

To construct a comprehensive vocabulary of rare diseases, we leverage the Orphanet Rare Disease Ontology (ORDO) \cite{weinreich2008orphanet}. ORDO provides a structured nomenclature and ontological representation of rare disease concepts, serving as a valuable resource for standardizing and organizing information related to rare diseases. We extract all disease concepts and their associated synonyms from ORDO to form our initial rare disease term list. This process ensures that we capture a wide range of rare disease names and their variations, establishing a solid foundation for our vocabulary.

However, rare diseases often suffer from name variations and inconsistent terminology usage \cite{shen2014entity}, which can hinder effective identification and extraction from clinical text. This challenge arises due to the complex nature of rare diseases, the evolving understanding of their underlying mechanisms, and the lack of consensus in naming conventions. As a result, the same rare disease may be referred to by multiple names or acronyms across different sources, making it difficult to develop a comprehensive and unified vocabulary.

To address this challenge and enhance the coverage of our rare disease vocabulary, we utilize the Unified Medical Language System (UMLS) as an intermediary dictionary \cite{bodenreider2004unified}. UMLS is a metathesaurus that integrates numerous biomedical vocabularies, including standard disease terminologies such as SNOMED CT, ICD-10, and MeSH. By mapping ORDO disease concepts to their corresponding UMLS concept unique identifiers (CUIs), we leverage the extensive synonymy information available in UMLS to expand our rare disease term coverage.

The mapping process involves several steps. First, we use string matching techniques to identify potential UMLS concepts that align with the ORDO disease concepts. This initial mapping is then refined using semantic type information and hierarchical relationships within UMLS to ensure the accuracy and specificity of the mapped concepts. By incorporating the synonyms and alternate names associated with each mapped UMLS concept, we substantially increase the breadth and depth of our rare disease vocabulary.

This comprehensive mapping process results in a final vocabulary consisting of 4,064 rare disease phenotype mappings from ORDO to UMLS. This expanded vocabulary not only includes the primary names of rare diseases but also encompasses a wide range of synonyms, acronyms, and alternate terms, thereby enhancing our ability to identify rare diseases mentioned in clinical text data.
To further align our rare disease phenotypes with the diagnostic coding systems used in the dataset for this study, we map the identified phenotypes to their corresponding ICD-9 \cite{world1988international} and ICD-10 \cite{world2004international} codes. This additional mapping step is crucial because patients in the dataset are coded using either or both of these classification systems. By establishing a link between our rare disease vocabulary and the standard diagnostic codes, we facilitate the integration of our findings with existing clinical workflows and enable seamless comparison with structured diagnostic information.

The ICD-9 and ICD-10 mapping process involves leveraging the inherent relationships between UMLS concepts and these coding systems. Many UMLS concepts are already associated with their corresponding ICD codes, allowing for a direct mapping. In cases where a direct mapping is not available, we employ a combination of automated and manual techniques to establish the appropriate connections. This includes utilizing existing mapping resources, such as the UMLS ICD-9/10 mappings, as well as consulting domain experts to validate and refine the mappings for specific rare disease phenotypes.

\subsection{Dictionary-Based Text Phenotyping}

To initiate the process of extracting relevant clinical entities from unstructured electronic health records (EHRs), we employ SemEHR \cite{wu2018semehr}, a state-of-the-art dictionary-based natural language processing (NLP) tool. SemEHR has demonstrated exceptional performance in extracting and normalizing a wide range of clinical concepts, including diseases, medications, and procedures, making it an ideal choice for our rare disease identification pipeline.

One of the key strengths of SemEHR lies in its ability to effectively handle the complexities and variability of clinical language. It employs advanced techniques such as named entity recognition (NER) and entity linking (EL) to accurately identify and extract relevant clinical information from unstructured text. NER focuses on recognizing and classifying named entities, such as diseases, drugs, and anatomical terms, within the clinical narrative. 

Once the named entities are identified, SemEHR performs entity linking to map the extracted entities to standardized terminologies, such as the Unified Medical Language System (UMLS). This mapping process involves disambiguating the extracted entities and linking them to their corresponding concept unique identifiers (CUIs) in the UMLS metathesaurus. By leveraging the rich semantic network and hierarchical relationships within UMLS, SemEHR can normalize the extracted entities to a common representation, enabling consistent and standardized analysis across different EHR systems and data sources.

To ensure that the mentions extracted by SemEHR are relevant to rare diseases, we employ the rare disease concept mappings as a filtering mechanism. By cross-referencing the extracted mentions with the rare disease concept mappings derived in the previous step, we can effectively identify and retain only those entities that are pertinent to rare diseases.

\subsection{LLM-Based Text Phenotyping}

After obtaining the initial results from SemEHR, we observe that the system produces a significant number of false positive extractions. This issue can be attributed to two main factors. First, the dictionary-based approach often struggles with the appropriate extraction of clinical abbreviations. For instance, the abbreviation \revise{``}PID" can refer to \revise{``}Primary Immunodeficiency," which is a rare disorder, but it can also be used to denote \revise{``}Pelvic Inflammatory Disease," a more common condition. Without considering the context in which the abbreviation appears, SemEHR may incorrectly identify it as a rare disease mention.

Second, SemEHR occasionally extracts rare disease mentions that are expressed in a hypothetical or negated context. For example, a clinical note might state, \revise{``}the patient does not have Huntington's disease," indicating the absence of the condition. However, SemEHR may still identify \revise{``}Huntington's disease" as a positive mention, leading to a false positive extraction. These contextual nuances are challenging for dictionary-based systems to handle, as they primarily rely on pattern matching and lack the ability to understand the surrounding linguistic context.

To address these limitations and improve the accuracy of rare disease phenotype extraction, we propose exploring the use of large language models (LLMs) with contextual reasoning capabilities. LLMs, such as LLaMA \cite{touvron2023llama}, which has demonstrated remarkable performance in natural language understanding tasks, particularly in capturing the semantic meaning and contextual information within text. By leveraging the advanced linguistic knowledge and reasoning abilities of these models, we aim to reduce the number of false positive identifications and enhance the overall quality of rare disease phenotype extraction.

The task of contextual reasoning for rare disease mention classification can be formulated as a binary classification problem. Given a rare disease mention $m$ extracted by SemEHR and its surrounding context information $c$ from the clinical text $T$, the LLM aims to predict the label $y$:
$$
y = f(m, c) \in {0, 1}
$$
where $y$ represents the predicted label for the rare disease mention. A label of $y = 1$ indicates a true positive, meaning that the mention refers to the presence of the rare disease in the given context. Conversely, a label of $y = 0$ indicates a false positive, suggesting that the mention does not actually indicate the presence of the rare disease based on the contextual information.

The function $f$ represents the LLM's classification mechanism, which maps the mention $m$ and its context $c$ to a binary label. This function encapsulates the model's ability to understand and reason about the linguistic context surrounding the mention, enabling it to make informed predictions about the mention's validity. This contextual reasoning step is crucial for improving the accuracy of rare disease phenotype extraction. By analyzing the surrounding context $c$, the LLM can filter out false positives caused by ambiguities in clinical language, such as abbreviations, negations, and other complexities.

\section{Experiments Design}
\subsection{Dataset}

\begin{figure}[h]
  \includegraphics[width=\columnwidth]{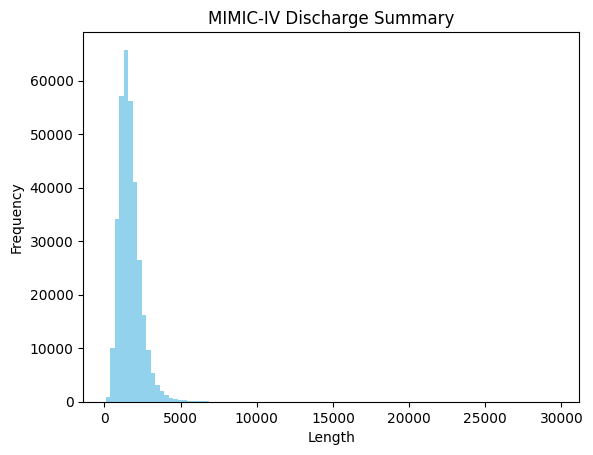}
  \caption{Data Distribution of MIMIC-IV's discharge summary lengths.}
  \label{fig:hist}
\end{figure}

For this study, we use the real-world free-text EHR data, MIMIC-IV. Our study mainly focuses on the discharge summaries from this database, which contains 331,794 reports from 145,915 patients \cite{johnson2023mimic}. Discharge summaries are a critical component of the electronic health record, as they provide a detailed recap of a patient's hospital stay, including the reasons for admission, the treatments and procedures performed, the patient's progress, and the final diagnosis and recommendations for ongoing care. The narrative structure and detailed clinical descriptions in discharge summaries can offer valuable contextual cues to identify rare disease patients. Figure 1 illustrates the highly skewed distribution of report lengths, with an average of 1,669 words, ranging from a minimum of 87 to a maximum of 29,684 words, which may provide the challenge of long contextual understanding (shown in Fig \ref{fig:hist}).

\subsection{Data Annotation}

To evaluate the performance of our rare disease phenotype identification approach, we create a gold standard dataset by randomly selecting 200 discharge summaries from the MIMIC-IV database. 
Two domain experts are tasked with manually annotating each mention of a rare disease in these summaries, classifying them as either a true positive (indicating the patient indeed has the rare disease) or a false positive (the mention does not actually indicate the presence of the rare disease). To ensure consistency and accuracy in the annotation process, the annotators are provided with a detailed guideline that includes specific examples on how to handle various scenarios, such as hypothetical or negated mentions. This guideline helps to standardize the annotation process and minimize subjectivity.

Based on the annotations provided by the two experts, we calculate the inter-annotator agreement using Cohen's kappa, a widely used statistical measure that assesses the level of agreement between two raters while accounting for chance agreement. The resulting Cohen's kappa score of 0.77 indicates a substantial level of agreement between the annotators, suggesting that the annotations are reliable and consistent.
To further enhance the quality of the gold standard dataset, any disagreements between the two annotators are resolved by a third annotator. This third annotator reviews the mentions where the initial annotators had differing opinions and makes a final decision on the classification of those mentions. 

In total, the gold standard dataset comprises 362 rare disease related mentions identified from the 200 discharge summaries. The gold standard dataset serves as a valuable resource for assessing the accuracy of our contextual rare disease phenotype identification model. By comparing the model's predictions against the expert-annotated labels, we can measure its performance in distinguishing true positive rare disease mentions from false positives in real-world clinical text. 

\subsection{Experiment Setup}

\subsubsection{Baseline}
\revise{Our methodology begins with the implementation of SemEHR as our baseline approach for extracting ontology-based rare diseases. We leverage SemEHR's capability to identify and normalize clinical concepts using predefined dictionaries and rules, which provides an initial set of potential rare disease mentions from the clinical text.}

\revise{To provide a comprehensive evaluation, we also compare our approach with several state-of-the-art models, categorized as follows:}

\paragraph{\revise{BERT-based models:}}
\revise{As these models are not specifically fine-tuned for rare diseases, we employ a two-step process. First, we utilize them for medical Named Entity Recognition (NER), followed by a dictionary filtering step using the rare disease ontology described in Section~\ref{section:RD_onto}. The models evaluated are:}

\begin{enumerate}
    \item \revise{PhenoBERT~\cite{feng2022phenobert}: A BERT-based model fine-tuned for disease phenotyping.}
    \item \revise{BioBERT-NER~\cite{alonso2021named}: Another BERT-based model fine-tuned for biomedical named entity recognition and normalization of diseases.}
\end{enumerate}

\paragraph{\revise{Large Language Model (LLM)-based approaches:} }

\revise{We also evaluate two advanced LLM-based models specifically designed for phenotype and rare disease recognition:}

\begin{enumerate}
    \item \revise{PhenoGPT~\cite{yang2024enhancing}: A fine-tuned model for phenotype recognition. As recommended, we utilize their LLaMA2-based version for this comparison.}
    \item \revise{PhenoHPO~\cite{wang2024fine}: A LLaMA2-based model fine-tuned for rare disease concept recognition and normalization.}
\end{enumerate}

\subsubsection{LLM}

Our proposed approach also includes leveraging large language models (LLMs) to perform further contextual reasoning and filter out negative mentions from SemEHR. 

For LLM selection, we focus on models within the 8 billion parameter range due to computational resource limitations, which often resemble resource-constrained clinical environments. This choice ensures that our approach remains feasible and practical for real-world clinical settings. We choose three state-of-the-art LLMs: \textbf{LLaMA3-8B}\footnote{https://huggingface.co/meta-llama/Meta-Llama-3-8B}, \textbf{Mistral-7B}\footnote{https://huggingface.co/mistralai/Mistral-7B-Instruct-v0.2}, and \textbf{Phi3-mini}\footnote{https://huggingface.co/microsoft/Phi-3-mini-128k-instruct}. These models have shown strong performance in various natural language processing tasks and provide a diverse set of architectures and training approaches.
To compare the performance between general domain LLMs and medical fine-tuned LLMs, we also select three medical LLMs: \textbf{OpenBioLLM}\footnote{https://huggingface.co/aaditya/Llama3-OpenBioLLM-8B}, \textbf{BioMistral}\footnote{https://huggingface.co/BioMistral/BioMistral-7B}, and \textbf{Alpacare}\footnote{https://huggingface.co/xz97/AlpaCare-llama2-7b}. These medical LLMs have been specifically fine-tuned on biomedical and clinical texts, and their inclusion allows us to assess the impact of domain-specific knowledge on the rare disease identification task.
OpenBioLLM-8B builds upon the latest LLaMA3-8B model and incorporates the DPO dataset and fine-tuning recipe along with a custom diverse medical instruction dataset. This fine-tuning process adapts the general-purpose LLM to the biomedical domain, potentially improving its performance on rare disease identification. Alpaca builds on LLaMA2-7B and is tuned on medical instructions, providing another perspective on the effectiveness of domain-specific fine-tuning. BioMistral utilizes Mistral-7B as its foundation model and is further pre-trained on PubMed Central, a large corpus of biomedical literature, which may enhance its ability to capture rare disease-related information.

To ensure robust and stable responses from the LLMs, we set the temperature parameter to 0 during inference. This setting reduces the variability in the generated outputs and promotes more deterministic behavior, which is desirable for the rare disease identification task.

In terms of prompt engineering, we experiment with both zero-shot and few-shot prompting, as well as knowledge-augmented generation. Zero-shot prompting involves providing the LLMs with a task description without any examples, relying on their inherent knowledge and understanding to generate appropriate responses. Few-shot prompting, on the other hand, includes a small number of exemplary rare disease mentions and their corresponding labels to guide the LLMs' predictions. Knowledge-augmented generation involves incorporating additional domain-specific information, such as rare disease definitions or phenotypic characteristics, into the prompts to enhance the LLMs' contextual understanding.

To investigate the impact of contextual information on the LLMs' ability to accurately identify rare diseases, we vary the amount of surrounding text provided to the models. We start with the full discharge summary paragraph as the input context, allowing the LLMs to consider the entire narrative when making predictions. However, processing the full paragraph may be computationally expensive and potentially introduce noise. Therefore, we gradually reduce the context to shorter lengths, such as a few sentences or a fixed window size around the rare disease mention.

By systematically evaluating the LLMs' performance across different context window sizes, we aim to identify the optimal balance between computational efficiency and accuracy. This analysis helps us determine the minimum amount of contextual information required for the LLMs to make accurate predictions, considering the trade-off between model performance and computational resources in resource-constrained clinical settings.

Through this comprehensive evaluation, we assess the effectiveness of different LLMs, prompting strategies, and context window sizes for the rare disease identification task. The use of prompt templates can be found in Additional file 1. By comparing the performance of general domain LLMs and medical fine-tuned LLMs, we gain insights into the impact of domain-specific knowledge on the task. The exploration of zero-shot, few-shot, and knowledge-augmented prompting allows us to identify the most effective approach for eliciting accurate responses from the LLMs. Furthermore, the analysis of context window sizes provides valuable information on the optimal balance between contextual information and computational efficiency.

The findings from this evaluation will inform the development of an optimized rare disease identification pipeline that leverages the strengths of both SemEHR and LLMs. By combining the initial extraction capabilities of SemEHR with the contextual reasoning abilities of LLMs, we aim to achieve high accuracy in identifying rare diseases from clinical texts while considering the practical constraints of resource-limited clinical environments. This hybrid approach has the potential to significantly improve the efficiency and effectiveness of rare disease identification, ultimately benefiting patient care and research in the field of rare diseases.

\section{Results}

\begin{table}[]
\begin{tabular}{lllll}
\hline
           & Size              &  F1 &Precision       & Recall          \\ \hline
SemEHR    &      &0.4866          & 0.3415          & 0.8458          \\ \hline
\multicolumn{4}{l}{\textit{Zero-Shot}}                                    \\ \hline
 \makebox[3mm]{}+AlpaCare   &  7B   &0.5083          & 0.4083          & 0.6732          \\
 \makebox[3mm]{}+BioMistral &   7B  &0.5219          & 0.4151          & 0.7028          \\
 \makebox[3mm]{}+OpenBioLLM & 8B   & 0.5568          & 0.4444          & 0.7453          \\
 \makebox[3mm]{}+Mistral    & 7B   & 0.5743          & 0.4949          & 0.6841          \\
 \makebox[3mm]{}+LLaMA3     &   8B  &0.6834          & 0.6023          & \textbf{0.7897} \\
 \makebox[3mm]{}+Phi3-mini  &   3.8B  &\textbf{0.6921} & \textbf{0.6197} & 0.7836          \\ \hline
\multicolumn{4}{l}{\textit{Few-Shot}}                                     \\ \hline
 \makebox[3mm]{}+AlpaCare   &   7B &0.5448          & 0.4839          & 0.6232         \\
 \makebox[3mm]{}+Mistral    &  7B   &0.5891          & 0.5497          & 0.6346          \\
 \makebox[3mm]{}+BioMistral &  7B   &0.5896          & 0.5110          & 0.6968          \\
\makebox[3mm]{}+OpenBioLLM &   8B  &0.6726          & 0.6908          & 0.6553          \\
 \makebox[3mm]{}+Phi3-mini  &   3.8B  &0.7484          & 0.6900          & \textbf{0.8176} \\
 \makebox[3mm]{}+LLaMA3     &  8B   &\textbf{0.7492} & \textbf{0.7071} & 0.7966          \\ \hline
 \multicolumn{4}{l}{\textit{KAG}}                                     \\ \hline
 \makebox[3mm]{}+AlpaCare   &   7B &0.5006          & 0.3964          & 0.6791         \\
 \makebox[3mm]{}+BioMistral &  7B   &0.5296          & 0.4222          & 0.7103          \\
 \makebox[3mm]{}+OpenBioLLM &   8B  &0.5547          & 0.4426          & 0.7427   \\
  \makebox[3mm]{}+Mistral    &  7B   &0.5647          & 0.4794          & 0.6871          \\
 \makebox[3mm]{}+LLaMA3     &  8B   &0.6830 & 0.6011 & 0.7907          \\
  \makebox[3mm]{}+Phi3-mini  &   3.8B  &\textbf{0.6966}          & \textbf{0.6221}          & \textbf{0.7913} \\
  \hline
\end{tabular}
\caption{\label{tab:results}
    Overall model performance for rare disease identification. B: Billion. }
\end{table}

\paragraph{Overall Performance.}

\revise{The results presented in Table \ref{tab:results} and \ref{tab:bert} demonstrate varying performance across different models for rare disease identification. Each model exhibits unique strengths and limitations, highlighting the complexity of this task.}

\revise{SemEHR, our baseline model utilizing a dictionary-based approach, achieves an F1 score of 0.4866. Its standout feature is a notably high recall of 0.8458, significantly outperforming other models in this metric. This high recall indicates that SemEHR is particularly adept at comprehensively capturing entities within the text, suggesting a superior ability to identify a wide range of relevant information.}

\revise{However, SemEHR's precision (0.3415) is comparatively lower, indicating a substantial number of false positive predictions. This suggests that many of the rare disease phenotypes identified by SemEHR are not actually present in the clinical text, highlighting the limitations of solely relying on dictionary-based methods for accurate rare disease identification.}

To address the limitations of the dictionary-based approach and further refine the results obtained from SemEHR, we employed various LLMs as an additional filtering step. Table \ref{tab:results} presents a comprehensive comparison of our proposed approach. The baseline model, SemEHR, which relies on a dictionary-based approach, achieves an F1 score of 0.4866. It achieves a high recall of 0.8458. However, its low precision score of 0.3415 indicates a substantial number of false positive predictions, suggesting that many of the rare disease phenotypes identified by SemEHR are not actually present in the clinical text. This highlights the limitations of solely relying on dictionary-based methods for accurate rare disease identification.

When using zero-shot prompting, Phi3-mini demonstrates the best performance, achieving an F1 score of 0.6921, closely followed by LLaMA3. These results suggest that LLMs can effectively leverage their pre-trained knowledge to identify rare diseases in clinical text without the need for task-specific fine-tuning. The improved performance of these models compared to the baseline SemEHR emphasizes the potential of LLMs in enhancing the accuracy of rare disease identification.

Interestingly, AlpaCare exhibits the least improvement in precision among the LLMs evaluated, indicating that the model may struggle to effectively discriminate between true and false positive rare disease mentions. This observation underscores the importance of carefully selecting and evaluating LLMs for the specific task of rare disease identification, as their performance can vary significantly.

\revise{Further, we also compare our hybrid method with other models for another baseline comparisons. As shown in Table \ref{tab:bert}, BioBERT-NER and PhenoBERT demonstrate similar performance profiles. Their precision scores (0.3145 and 0.3465 respectively) are comparable to SemEHR. However, their recall scores (0.2928 and 0.3750) are substantially lower. This suggests that while these BERT-based models maintain a similar level of accuracy in the entities they do identify, they are more prone to missing relevant entities compared to SemEHR. The lower recall implies a higher rate of false negatives, indicating a more conservative approach to entity recognition.}

\revise{The two LLM fine-tuning models, PhenoGPT and PhenoHPO show better performance than BERT based models and SemEHR. PhenoGPT shows the best F1 score among the five models excluding our method, coupled with a strong precision (0.4673), suggesting a well-balanced performance in both accurately identifying entities and capturing a good proportion of them. This indicates the potential of Large Language Models (LLMs) in enhancing the accuracy of rare disease identification. Despite being fine-tuned with rare disease phenotyping data, PhenoGPT and PhenoHPO demonstrate marginal performance improvements, highlighting the persistent challenges in this domain. Finally, our hybrid method achieves the best performance, evidenced by a significantly higher F1 score compared to traditional approaches. Moreover, these methods achieve a more balanced precision-recall trade-off, underscoring their robustness and applicability across diverse scenarios.}

\begin{table}[h]
\begin{tabular}{lllll}
\hline
          & \revise{F1}     & \revise{Precision} & \revise{Recall} & \revise{Approach}                                    \\ \hline
\revise{BioBERT-NER \cite{alonso2021named}}   & \revise{0.3032} & \revise{0.3145}    & \revise{0.2928} & \revise{BERT-based \& dictionary filtering }    \\
\revise{PhenoBERT \cite{feng2022phenobert}} & \revise{0.3020}  & \revise{0.3465}    & \revise{0.3750}  & \revise{BERT-based \& dictionary filtering }    \\
\revise{SemEHR \cite{wu2018semehr}}    & \revise{0.4866} & \revise{0.3415}    & \revise{0.8458} & \revise{Semantic-based \& dictionary filtering} \\ 
\revise{PhenoGPT \cite{yang2024enhancing}} & \revise{0.5483} & \revise{0.4673}    & \revise{0.6632} & \revise{LLM fine-tuning}                            \\
\revise{PhenoHPO \cite{wang2024fine}} & \revise{0.3945} & \revise{0.3589}    & \revise{0.4379} & \revise{LLM fine-tuning}                            \\
\revise{Ours*}     & \revise{\textbf{0.7492}} & \revise{\textbf{0.7071}}    & \revise{\textbf{0.7966}} & \revise{Dictionary filtering \& LLM prompting} \\ \hline
\end{tabular}
\caption{\label{tab:bert}
    \revise{Comparison of models for rare disease identification from texts. *We use the best performance in Table \ref{tab:results} as our method here, which is the combination of SemEHR and LLaMA3 (with few-shot).} }
\end{table}

The overall results demonstrate the effectiveness of combining dictionary-based methods with LLMs for improved rare disease identification. By leveraging the strengths of both approaches -- the comprehensive coverage of dictionary-based methods and the contextual understanding of LLMs -- we can achieve a more accurate and reliable identification of rare diseases in clinical text. This hybrid approach holds promise for facilitating the early detection and management of rare diseases, ultimately leading to better patient outcomes.

\paragraph{Few-Shot Prompting.}

Few-shot prompting is a promising approach for enhancing the performance of LLMs in rare disease identification tasks. To investigate the impact of few-shot prompting, we conduct experiments with varying numbers of shots, ranging from 1 to 10. The examples used for few-shot prompting are randomly selected, ensuring no overlap with the test dataset to maintain the integrity of the evaluation. Here is the template used for few-shot prompting:

\begin{lstlisting}[style=promptstyle]
[Task description]: 
Given a text, the start and end indices of a mention from the text, and the mention itself, determine whether the mention is a true rare disease mention or not.
[Instructions]: 
Return "yes" if it is a true mention, "no" if not a true mention. Please only answer yes or no in your response. 
[Report]:
REPORT
[Mention]:
Amyotrophic Lateral Sclerosis
[Start and end index]:
(111, 140)
[Few-shot examples]: 
Example 1. REPORT. MENTION. (START, END). YES/NO
Example 2. REPORT. MENTION. (START, END). YES/NO
...
[Answer]
\end{lstlisting}

Table \ref{tab:results} reports the best performance achieved by each model using few-shot prompting. Notably, all models demonstrate improvements in F1 score compared to their zero-shot counterparts, with increases ranging from 0.01 to 0.12. OpenBioLLM exhibits the most substantial improvement, with an impressive 0.1158 increase in F1 score. This significant improvement suggests that OpenBioLLM effectively leverages the additional information provided by the few-shot examples to refine its predictions and better identify rare diseases in clinical text. On the other hand, Mistral shows the least improvement, with a modest 0.0148 increment in F1 score, indicating that the model may have a limited ability to capitalize on the few-shot examples for this specific task.

\begin{figure}[h]
\centering
  \includegraphics[width=1\columnwidth,trim={60 0 0 0},clip]{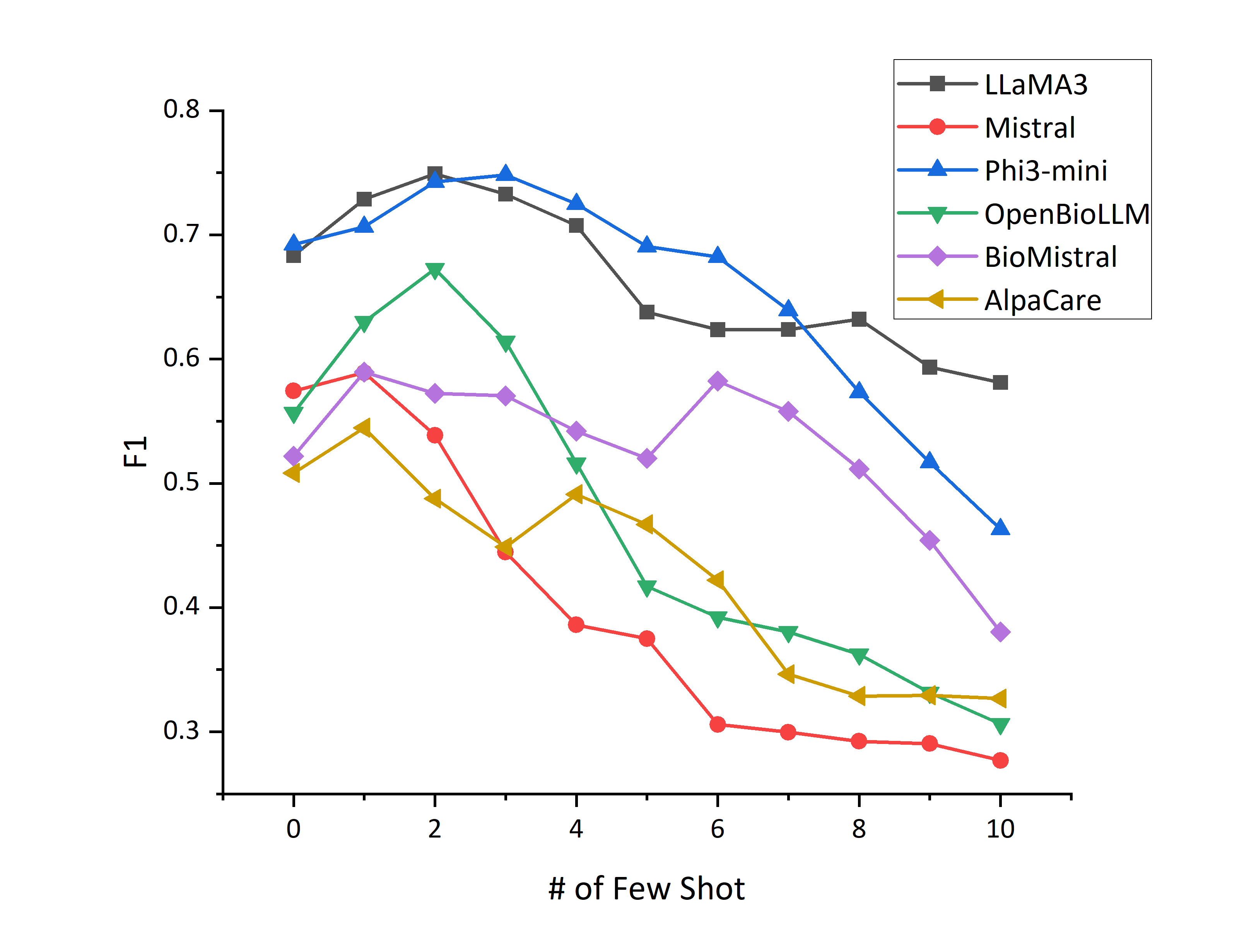}
  \caption{Few-shot prompting performance (F1) of LLMs.}
  \label{fig:fewshot}
\end{figure}

Figure \ref{fig:fewshot} provides a more comprehensive analysis of the impact of the number of few-shot examples on the performance of different LLMs. The plot reveals an intriguing trend: increasing the number of few-shot examples initially leads to improved performance for all LLMs, as evidenced by the upward trend in F1 scores when adding 1-3 shots. This observation suggests that providing a small set of representative examples can effectively guide the LLMs to better understand the task and capture the relevant patterns associated with rare disease mentions.

\revise{However, the results also highlight an important consideration: more examples do not necessarily guarantee better performance. As the number of few-shot examples increases beyond a certain point, the F1 scores begin to decline, indicating that an excessive number of examples can actually hinder the models' ability to accurately identify rare diseases in clinical text. This finding emphasizes the importance of striking the right balance in the number of few-shot examples to optimize the effectiveness of few-shot learning for rare disease identification.
Among the LLMs evaluated, LLaMA3 consistently demonstrates the best performance across different numbers of few-shot examples, closely followed by Mistral and Phi3-mini. These models exhibit a steady improvement in F1 score as the number of few-shot examples increases, showcasing their strong capacity to leverage the additional information provided by the examples to refine their predictions. This observation highlights the potential of these LLMs to effectively adapt to the specific task of rare disease identification through few-shot learning.}

In conclusion, few-shot prompting proves to be a valuable technique for enhancing the performance of LLMs in rare disease identification. By providing a small set of representative examples, LLMs can better grasp the nuances of the task and improve their ability to accurately identify rare diseases in clinical text. However, it is crucial to find the optimal balance in the number of few-shot examples to maximize the effectiveness of this approach. The results also underscore the varying capabilities of different LLMs in leveraging few-shot examples, with LLaMA3, Mistral, and Phi3-mini demonstrating particularly strong performance in this regard.

\paragraph{Knowledge Augmented Generation.}

To further explore the potential of enhancing the performance of LLMs in rare disease identification, we investigate the impact of incorporating external knowledge into the prompts through Knowledge Augmented Generation (KAG). Specifically, for each rare disease mention encountered in the clinical text, we extract its corresponding definition from the Unified Medical Language System (UMLS) and integrate it into the prompts. \revise{The prompt template can be found in the additional files.} By providing this additional contextual information, we aim to evaluate whether the LLMs can effectively leverage this new knowledge alongside their pre-existing training data to generate improved responses and accurately identify rare diseases.

Here is the template used for KAG prompts:

\begin{lstlisting}[style=promptstyle]
[Task description]: 
Given a text, the start and end indices of a mention from the text, and the mention itself, determine whether the mention is a true rare disease mention or not.
[Instructions]: 
Return "yes" if it is a true mention, "no" if not a true mention. Please only answer yes or no in your response. 
[Report]:
REPORT
[Mention]:
Amyotrophic Lateral Sclerosis
[Start and end index]:
(111, 140)
[Definition of Amyotrophic Lateral Sclerosis]: 
A neurodegenerative disease characterized by progressive muscular paralysis reflecting degeneration of motor neurons in the primary motor cortex, corticospinal tracts, brainstem and spinal cord. 
[Answer]
\end{lstlisting}

Our findings reveal mixed results regarding the effectiveness of KAG in this task. Notably, the Phi3-mini and BioMistral models demonstrate marginally higher F1 scores when compared to their zero-shot counterparts. This slight improvement suggests that these models are capable of incorporating the external knowledge provided by UMLS definitions to some extent, leading to a modest enhancement in their ability to identify rare diseases accurately. However, it is important to note that the performance of these models in the KAG setting still falls short of their few-shot prompting results, indicating that the incorporation of external knowledge alone may not be as effective as providing task-specific examples for guiding the models' predictions.

Interestingly, the other LLMs evaluated in this study fail to benefit from the incorporation of knowledge into the prompts, as evidenced by the lack of improvement in their F1 scores compared to the zero-shot setting. \revise{This observation suggests that these models may already possess sufficient intrinsic knowledge about rare diseases, rendering the additional information provided by UMLS definitions redundant or less impactful.} It is plausible that the extensive pre-training of these LLMs on large-scale biomedical and clinical text corpora has equipped them with a comprehensive understanding of rare diseases, making the external knowledge less crucial for their performance in this specific task. \revise{Another reason can be that UMLS definitions are often broad and may not capture the specific nuances of how rare diseases present in clinical text. Also, the formal language of UMLS definitions might not align well with the more varied and informal language used in clinical notes.}

Moreover, the superior performance of few-shot prompting compared to KAG hints at the importance of task-specific examples in facilitating the models' comprehension and adaptation to the rare disease identification task. By providing a small set of representative examples, few-shot prompting allows the LLMs to grasp the nuances and patterns associated with rare disease mentions more effectively than the incorporation of general definitional knowledge. This finding highlights the significance of carefully curating task-specific examples to guide the models' learning process and optimize their performance in specialized domains like rare disease identification.

In conclusion, while Knowledge Augmented Generation shows promise in enhancing the performance of LLMs by incorporating external knowledge, its effectiveness in the task of rare disease identification appears to be limited. The marginal improvements observed in the Phi3-mini and BioMistral models suggest that these LLMs can benefit from the integration of UMLS definitions to a certain extent. However, the overall results indicate that the models may already possess sufficient intrinsic knowledge about rare diseases, and that few-shot prompting, with its task-specific examples, proves to be a more effective approach for improving their performance in this domain. These findings underscore the importance of carefully considering the specific characteristics of the task and the pre-existing knowledge of the LLMs when designing strategies to enhance their performance in specialized applications like rare disease identification.

\paragraph{Medical LLMs vs General LLMs.}

One of the most intriguing findings of our study is the comparative performance of medical fine-tuned LLMs and general domain LLMs in the task of rare disease identification. Contrary to expectations, the medical fine-tuned LLMs, such as OpenBioLLM, BioMistral, and AlpaCare, demonstrate inferior performance compared to their general domain counterparts. This observation highlights the need for further research and development efforts to optimize the fine-tuning process and effectively incorporate medical knowledge into these specialized LLMs.

The suboptimal performance of medical fine-tuned LLMs can be attributed to several key challenges. One major issue is the lack of robustness and limited understanding of the given instructions and clinical reports exhibited by these models. Despite undergoing fine-tuning on medical corpora, these LLMs struggle to fully grasp the nuances and complexities of the clinical language and context, leading to subpar performance in identifying rare diseases accurately. This finding suggests that the current fine-tuning strategies employed for medical LLMs may not be sufficient to capture the intricate relationships and domain-specific knowledge required for this task.

Moreover, the superior performance of general domain LLMs in rare disease identification underscores the importance of carefully evaluating and adapting LLMs for specific medical tasks. While medical fine-tuning aims to equip LLMs with domain-specific knowledge, the results of our study indicate that generic fine-tuning alone may not guarantee optimal performance in specialized tasks like rare disease identification. This observation calls for a more nuanced approach to fine-tuning, taking into account the unique characteristics and requirements of each medical task.

To address these challenges and improve the performance of medical LLMs, further research is necessary to develop more sophisticated fine-tuning strategies that can effectively leverage medical knowledge. This may involve exploring novel techniques for incorporating domain-specific information, such as ontology-based fine-tuning or knowledge graph integration, to enhance the LLMs' understanding of medical concepts and their relationships. Additionally, the development of task-specific fine-tuning approaches tailored to rare disease identification could help bridge the performance gap between medical and general domain LLMs.

Furthermore, the evaluation and adaptation of LLMs in the medical domain should go beyond generic fine-tuning and focus on comprehensive testing across a range of clinical tasks and datasets. By thoroughly assessing the performance of LLMs in various medical scenarios, researchers can identify the strengths and weaknesses of different models and fine-tuning strategies, enabling the development of more robust and reliable LLMs for clinical applications.

In conclusion, the inferior performance of medical fine-tuned LLMs compared to general domain LLMs in rare disease identification highlights the need for further advancements in the fine-tuning process and the incorporation of medical knowledge. The lack of robustness and limited understanding exhibited by these specialized LLMs underscores the importance of developing more sophisticated fine-tuning strategies and conducting comprehensive evaluations to ensure optimal performance in clinical tasks. By addressing these challenges through targeted research efforts, we can unlock the full potential of LLMs in the medical domain and enhance their accuracy and reliability in critical applications like rare disease identification.

\begin{figure}[h]
\centering
\includegraphics[width=1\columnwidth,trim={60 0 0 0},clip]{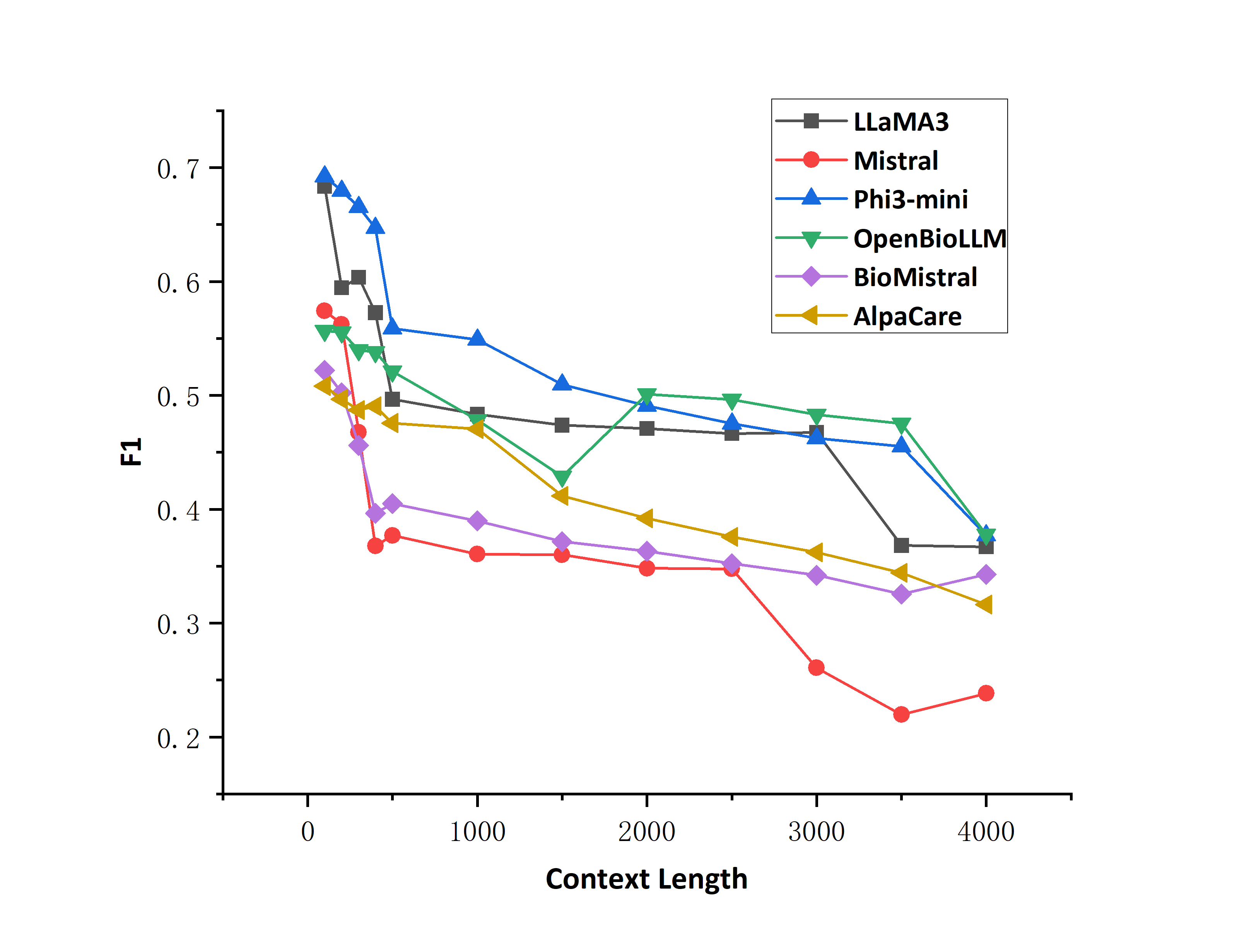}
  \caption{Impact of context length on the performance (F1) of LLMs.}
  \label{fig:window}
\end{figure}

\paragraph{The Impact of Context Length.}
Figure \ref{fig:window} illustrates the impact of context length on the performance of the LLMs in the task of rare disease identification. The context length is systematically varied from 100 to 4000 words, as some of the LLMs do not support input length larger than 4096. A key observation from the figure is that as the context length increases, the performance of the models generally decreases, as indicated by the downward trend in the plot. \revise{This was primarily due to the introduction of noise and irrelevant information in longer contexts, which can confuse the models.} This suggests that smaller context lengths may be more effective for LLMs to make accurate inferences about the presence of rare diseases in clinical text. The models appear to perform better when focusing on more localized context, such as the most relevant sections of the clinical report, rather than considering the entire document. \revise{However, it is important to note that there were indeed specific cases where longer context lengths proved advantageous. These cases typically involved complex clinical scenarios where the rare disease mention was dependent on information spread across a larger portion of the clinical note. For instance, in cases where the rare disease diagnosis was contingent on a combination of symptoms, family history, and test results described throughout the discharge summary, the longer context allowed the model to capture these relationships more effectively.} This finding highlights the importance of selecting an appropriate context window size to optimize the performance of LLMs in rare disease identification tasks and strikes a balance between providing sufficient contextual information and avoiding the inclusion of irrelevant or potentially confounding details.

Furthermore, the results indicate that the current state-of-the-art LLMs still have difficulties in reasoning long sequences. Despite their impressive capabilities in various natural language processing tasks, the contextual reasoning ability of these models is not yet robust enough to effectively handle the full length of clinical reports. This observation underscores the need for further research and development of LLMs to improve their ability to reason over longer contexts and capture relevant information from comprehensive clinical narratives.

\begin{figure*}[h]
  \includegraphics[width=\columnwidth]{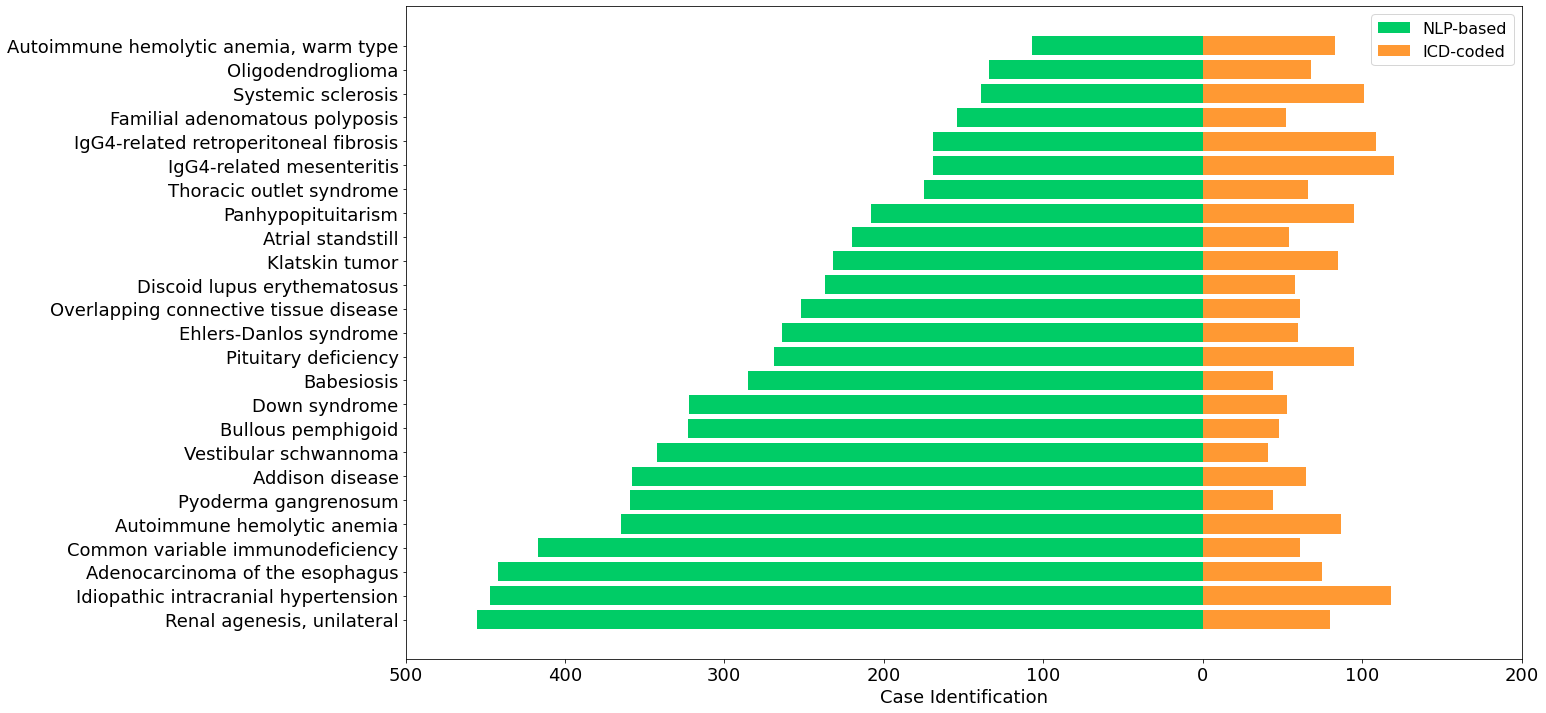}
  \caption{Case identification of rare diseases identified by NLP-based (free-text) and ICD-based (structured) data.}
  \label{fig:compare}
\end{figure*}

\paragraph{Analysis on the Full Dataset.}

To assess the real-world impact of our approach, we apply the best-performing method to the full dataset, consisting of over 331,000 discharge summaries from 145,915 patients. This comprehensive analysis reveals the presence of 1,143 distinct rare diseases within the cohort, affecting a total of 24,593 individuals. The scale of this analysis highlights the potential of our approach to identify a significant number of rare disease cases from a large, diverse patient population.

One of the most striking findings from this analysis is the discovery of 495 rare diseases that were not previously captured in the structured diagnostic data (i.e., ICD codes). These diseases were identified solely through the analysis of patients' discharge summaries using our NLP-based approach. This finding underscores the limitations of relying exclusively on structured diagnostic codes for rare disease identification and emphasizes the untapped potential of unstructured clinical narratives. By leveraging advanced NLP techniques, we can uncover previously unrecognized patients with rare diseases, potentially leading to improved diagnosis, treatment, and research opportunities for these individuals.

Furthermore, our analysis reveals that 337 rare diseases have a higher number of patients identified from free-text EHRs compared to structured diagnostic codes. This discrepancy suggests the presence of potentially undiagnosed or misdiagnosed individuals with rare diseases within the cohort. The higher prevalence of these diseases in the unstructured clinical narratives highlights the importance of comprehensive phenotyping approaches that go beyond traditional coding methods. By leveraging the rich information contained within free-text EHRs, we can identify patients who may have been overlooked or misclassified, enabling targeted interventions and specialized care for these individuals.

To illustrate the significance of these findings, Figure 5 presents a selection of rare diseases that were identified through our NLP-based analysis of free-text EHRs but were absent from the structured diagnostic codes. These examples serve as compelling evidence for the utility of applying advanced text mining techniques to unstructured clinical narratives. By uncovering these previously undetected rare diseases, we can gain valuable insights into their epidemiology, natural history, and potential treatment strategies. Moreover, the identification of these patients can facilitate their referral to specialized care centers and support their inclusion in relevant clinical trials and research initiatives.

The successful application of our approach to the full dataset demonstrates the power of combining large-scale clinical datasets with sophisticated NLP methods for rare disease identification. Our results are consistent with previous findings by Dong et al. [6] and Ford et al. [10], who have highlighted the value of incorporating unstructured clinical texts alongside structured coded data for disease identification. By leveraging the wealth of information contained within unstructured clinical narratives, we can enhance our understanding of rare diseases, identify patients who may have gone undiagnosed, and ultimately improve outcomes for this often-overlooked patient population.

The identification of a substantial number of previously unrecognized rare disease cases through our analysis underscores the potential impact of our approach on rare disease research and patient care. By uncovering these hidden cases, we can expand our knowledge of rare disease epidemiology, natural history, and potential therapeutic targets. Furthermore, the identification of these patients can facilitate their timely diagnosis, referral to specialized care, and inclusion in relevant research studies and clinical trials. This, in turn, can lead to improved outcomes, quality of life, and the development of targeted interventions for individuals with rare diseases.


\revise{In conclusion, our comprehensive analysis of the full dataset demonstrates the real-world applicability and impact of our NLP-based approach for rare disease identification. The discovery of a significant number of previously unrecognized rare disease cases highlights the limitations of relying solely on structured diagnostic codes and emphasizes the untapped potential of unstructured clinical narratives. This represents a significant advancement in clinical utility compared to traditional methods, offering improved case detection and more comprehensive phenotyping. By analyzing unstructured clinical narratives, we capture a more nuanced and complete picture of a patient's condition, including subtle symptoms and clinical observations often overlooked in structured data. This can support clinical decision-making by prompting clinicians to consider less common diagnoses. However, it is crucial to emphasize that our method is intended to complement, not replace, traditional diagnostic approaches, and the rare disease mentions identified should be viewed as signals for further clinical investigation rather than definitive diagnoses. By leveraging the power of large-scale clinical datasets and advanced NLP techniques, we can enhance our understanding of rare diseases, identify patients who may have gone undiagnosed, and ultimately improve outcomes for this often-overlooked patient population. These findings underscore the importance of integrating unstructured clinical texts with structured coded data for comprehensive rare disease identification and research.
}

\section{Conclusion}

In conclusion, this study introduces a novel hybrid approach that synergistically combines dictionary-based natural language processing (NLP) tools with large language models (LLMs) to enhance the identification of rare diseases from unstructured clinical reports. By capitalizing on the complementary strengths of these two techniques, the proposed method exhibits superior performance compared to traditional NLP systems and standalone LLMs. Our comprehensive experiments investigate various strategies, including zero-shot and few-shot prompting, knowledge-augmented generation (KAG), and the impact of context length on the model's performance. The results demonstrate that LLaMA3 and Phi3-mini consistently achieve the highest F1 scores in the task of rare disease identification.

Moreover, our analysis reveals the potential of the proposed approach to uncover previously unidentified rare disease cases that are not yet documented in structured medical records. This finding underscores the significant impact of our method in facilitating the early detection and diagnosis of rare diseases, which can ultimately lead to improved patient outcomes and the development of targeted treatment strategies.

However, it is important to acknowledge the limitations of our approach and the areas that require further advancements to ensure its optimal effectiveness and seamless integration into medical practice. One key factor influencing the performance of our method is the quality of ontology matching among the Unified Medical Language System (UMLS), the International Classification of Diseases, 10th Revision (ICD-10), and the Orphanet Rare Disease Ontology (ORDO). Inaccuracies in the ontology mappings, such as the example of \revise{``}Congenital pulmonary airway malformation" from UMLS being mapped to \revise{``}Q330 Congenital cystic lung" in ICD-10, can impact the precision of rare disease identification. \revise{To address these challenges, we have conducted manual reviews of complex mappings, seeking validation from domain experts, utilizing intermediate ontologies to improve alignment. Despite these efforts, we acknowledge that some level of imperfection in ontology mapping persists.} As medical terminologists continuously refine and update these mappings with the aid of ontology mapping tools, we anticipate an improvement in the method's performance over time.

Furthermore, our analysis primarily concentrates on comparing the case identification of rare diseases derived from NLP-based (free-text) and ICD-based (structured) data. It is crucial to recognize that there may be overlapping cases between these two approaches, and future research should focus on investigating the extent and implications of such overlap. By conducting a thorough analysis of the convergence and divergence between NLP-based and ICD-based rare disease identification, we can gain valuable insights into the complementary nature of these approaches and develop strategies to harmonize and integrate their findings.

Despite these limitations, our study establishes a solid foundation for the application of hybrid NLP and LLM approaches in the field of rare disease identification. The promising results obtained from our experiments highlight the immense potential of leveraging advanced language models and dictionary-based tools to unlock the wealth of information contained within unstructured clinical reports. By continually refining and expanding upon our methodology, we can work towards overcoming the challenges posed by ontology mapping inconsistencies and overlapping case identification, ultimately paving the way for more accurate and comprehensive rare disease detection.

In future research endeavors, it is essential to focus on enhancing the robustness and adaptability of our approach to accommodate the evolving nature of medical ontologies and terminologies. Collaborations between computational linguists, medical experts, and ontology developers will be crucial in addressing the limitations associated with ontology mapping and ensuring the seamless integration of our method into clinical practice. Additionally, exploring advanced techniques for data integration and harmonization, such as entity resolution and record linkage, can help to mitigate the impact of overlapping cases and provide a more holistic view of rare disease epidemiology.

Furthermore, the integration of our hybrid approach with other complementary methodologies, such as machine learning algorithms and knowledge graphs, can further enhance its performance and extend its applicability to a wider range of clinical scenarios. By leveraging the strengths of multiple techniques and data sources, we can develop a more comprehensive and robust framework for rare disease identification, ultimately improving patient care and accelerating research efforts in this critical domain.

\section*{Declarations}

\begin{itemize}
\item Funding \\ This work was supported by National Institute for Health and Care Research (NIHR202639), NIHR/HDR UK Winter Pressure Award (WP0006) and Medical Research Council (MR/S004149/2, MR/X030075/1). The views expressed are those of the author(s) and not necessarily those of the NIHR or the Department of Health and Social Care. This work was also supported by British Council (UCL-NMU-SEU International Collaboration On Artificial Intelligence In Medicine: Tackling Challenges Of Low Generalisability And Health Inequality; Facilitating Better Urology Care With Effective And Fair Use Of Artificial Intelligence - A Partnership Between UCL And Shanghai Jiao Tong University School Of Medicine). H.Wu’s role in this research were partially funded by the Legal \& General Group (research grant to establish the independent Advanced Care Research Centre at University of Edinburgh). The funders had no role in conduct of the study, interpretation, or the decision to submit for publication. The views expressed are those of the authors and not necessarily those of Legal \& General.

\item Competing interests\\
None declared
\item Ethics approval and consent to participate\\
This work used the de-identified clinical notes in MIMIC-IV. We completed the Collaborative Institutional Training Initiative (CITI) Program's “Data or Specimens Only Research” course\footnote{\url{https://physionet.org/content/shareclefehealth2013/view-required-trainings/1.0/\#1}} and signed the data use agreement to get access to the data. We strictly followed the guidelines and only used locally hosted LLMs with the data. 
\item Availability of data and materials \\
Data is available on Physionet with credentialed access: \url{https://www.physionet.org/content/mimic-iv-note/2.2/}
\item Author contribution \\
J.W., H.D., and H.Wu conceived the project. J.W. developed the methodology and experiments. Z.L., H.Wang and R.L. performed data annotation. J.w. and H.D. drafted manuscript. All authors contributed to the final manuscript preparation.

\item Acknowledgements\\The authors would like to thank all the reviewers and editors for their valuable feedback that substantially improved this paper.
\item Consent for publication\\ All authors consent for publication.
\end{itemize}

\bibliography{sn-bibliography}

\end{document}